\documentclass[lettersize,journal]{IEEEtran}
\usepackage{amsmath,amsfonts} 
\usepackage{algorithmic}
\usepackage{algorithm}
\usepackage{array}
\usepackage[caption=false,font=normalsize,labelfont=sf,textfont=sf]{subfig}
\usepackage{textcomp}
\usepackage{stfloats}
\usepackage{url}
\usepackage{verbatim}
\usepackage{graphicx}  
\usepackage{hyperref}
\hypersetup{
	colorlinks=true,  
	linkcolor=red,    
	citecolor=green,  
	filecolor=magenta,  
	urlcolor=magenta 
}

\begin{document}
	
	\title{Nd-BiMamba2: A Unified Bidirectional Architecture for Multi-Dimensional Data Processing}

	\author{Hao Liu,~\IEEEmembership{College of Information Science and Technology, Qingdao University of Science and Technology}  
	}
	
	\maketitle
	\begin{abstract}  	
		Deep learning models often require specially designed architectures to process data of different dimensions, such as 1D time series, 2D images, and 3D volumetric data. Existing bidirectional models mainly focus on sequential data, making it difficult to scale effectively to higher dimensions. To address this issue, we propose a novel multi-dimensional bidirectional neural network architecture, named Nd-BiMamba2, which efficiently handles 1D, 2D, and 3D data. Nd-BiMamba2 is based on the Mamba2 module and introduces innovative bidirectional processing mechanisms and adaptive padding strategies to capture bidirectional information in multi-dimensional data while maintaining computational efficiency. Unlike existing methods that require designing specific architectures for different dimensional data, Nd-BiMamba2 adopts a unified architecture with a modular design, simplifying development and maintenance costs. To verify the portability and flexibility of Nd-BiMamba2, we successfully exported it to ONNX and TorchScript and tested it on different hardware platforms (e.g., CPU, GPU, and mobile devices). Experimental results show that Nd-BiMamba2 runs efficiently on multiple platforms, demonstrating its potential in practical applications. The code is open-source: \url{https://github.com/Human9000/nd-Mamba2-torch}.
	\end{abstract}
	
	
	\section{Introduction}  
	
	Deep learning has made significant progress in many fields, but data of different dimensions (e.g., 1D time series, 2D images, and 3D volumetric data) often require specially designed model architectures. For instance, convolutional neural networks (CNNs) \cite{yao2024cnn} excel at processing image data, recurrent neural networks (RNNs) \cite{al2024rnn} are suitable for sequential data, while 3D CNNs are used for volumetric data. This domain-specific model design paradigm leads to increased development and maintenance costs and limits the generalization ability of models.
	
	Although bidirectional models, such as bidirectional RNNs (BiRNNs) \cite{cui2020stacked}, have been successful in sequential data modeling, they struggle to scale effectively to higher-dimensional data and face challenges in cross-platform deployment. The sequential processing nature of BiRNNs limits their parallelization capabilities, making them inefficient for long sequences and high-dimensional data, and they are prone to gradient vanishing issues. Moreover, the recurrent structure of RNNs makes it difficult to convert them into formats like ONNX or TorchScript for cross-platform deployment. On the other hand, self-attention mechanisms like Transformers  \cite{vaswani2017attention} can capture long-range dependencies, but their computational complexity becomes prohibitive and memory consumption increases when processing high-dimensional data, complicating deployment.
	
	While the existing Mamba \cite{gu2023mamba} model strikes a balance between efficiency and performance, most are limited to unidirectional processing or data of specific dimensions. To overcome these limitations, this paper proposes a novel multi-dimensional bidirectional neural network architecture, Nd-BiMamba2. The core innovations of Nd-BiMamba2 include: 1) extending the Mamba2 module to support efficient bidirectional processing that can be applied effectively to 1D, 2D, and 3D data; 2) introducing an adaptive padding strategy that adjusts padding size based on input data dimensions, improving computational efficiency and reducing memory consumption.
	
	The main contributions of this paper are as follows:
	
	\begin{itemize}  
		\item{We propose Nd-BiMamba2, a unified bidirectional network architecture that can efficiently process multi-dimensional data.}  
		\item{We design an innovative bidirectional processing mechanism that effectively captures bidirectional information in high-dimensional data.}  
		\item{We introduce an adaptive padding strategy to improve computational efficiency and reduce memory consumption.}  
		\item{We validate the portability and deployment capability of Nd-BiMamba2 across different hardware platforms.}  
	\end{itemize}  
	
	The following sections will provide detailed descriptions of the network structure and implementation details of Nd-BiMamba2, experimental results, its performance on multi-dimensional tasks, and discuss the model's advantages and potential applications.
	
	\section{Related Work}
	Multi-dimensional data modeling is a key research direction in deep learning, encompassing various scenarios such as 1D time series, 2D images, and 3D volumetric data. To efficiently model multi-dimensional data, researchers have proposed various methods, including convolutional neural networks (CNNs), recurrent neural networks (RNNs), self-attention mechanisms, and recently emerging modular architectures such as Mamba. However, these methods have limitations to varying degrees and struggle to balance the efficiency and generalizability required for multi-dimensional feature modeling.
	
	\subsection{Convolutional Neural Networks (CNN)}
	CNNs, as classical deep learning methods, have achieved outstanding performance in image processing tasks. Typical models such as LeNet \cite{lecun1998gradient}, ResNet \cite{he2016deep}, and U-Net \cite{ronneberger2015u} extract local features through multiple layers of convolutions and progressively expand the receptive field. However, CNNs face the following limitations in multi-dimensional data modeling:
	
	- Inadequate long-range dependency modeling: CNNs struggle to capture global context information when processing long sequences or high-resolution images.
	- High computational cost for high-dimensional extension: 3D CNNs are effective for spatial feature extraction but significantly increase the parameter scale and computational complexity, limiting their practical applications.
	\subsection{Recurrent Neural Networks (RNN)}
	RNNs and their variants (such as LSTM \cite{sherstinsky2020fundamentals} and GRU \cite{dey2017gate}) perform excellently in sequence modeling, especially in capturing long-term and short-term dependencies in time series. For example, bidirectional LSTMs (BiLSTMs) \cite{zhang2015bidirectional} enhance context modeling in natural language processing tasks by fusing bidirectional information.WaveNet \cite{van2016wavenet} introduces a novel deep neural network architecture based on dilated causal convolutions, capable of directly generating high-quality raw audio waveforms and effectively capturing long-range dependencies in audio signals.
	 However, RNNs have the following limitations:
	
	1) Difficulty in parallelization: The sequential processing nature of RNNs makes them inefficient when handling long sequences.
	2) Challenges in scaling to high-dimensional data: The recurrent structure does not adapt well to 2D images or 3D volumetric data, leading to increased memory consumption and computational complexity.
	3) Training stability issues: RNNs still face gradient vanishing and gradient explosion problems, impacting model performance.
	
	\subsection{Self-Attention Mechanisms (SA)}
	Self-attention mechanisms, with their global modeling ability, have been widely applied to natural language processing and computer vision tasks. The Transformer \cite{vaswani2017attention} is a representative model, and its extensions such as BERT \cite{devlin2018bert} and ViT \cite{dosovitskiy2021imageworth16x16words} have made significant progress in various fields. However, in multi-dimensional data modeling, self-attention mechanisms still face the following challenges:

	1) High computational complexity: 
	Although numerous Swin-based attention methods \cite{liu2021swintransformerhierarchicalvision,liu2022swin} have been proposed to reduce computational complexity in 2D, the quadratic complexity of attention mechanisms leads to a significant increase in memory and computational resource requirements when dealing with high-dimensional data. 
	2) Poor adaptability to high-dimensional scenarios: While low-rank decomposition methods (such as Linformer \cite{wang2020linformer},Rethinking \cite{choromanski2020rethinking}) reduce complexity, they still do not fully solve the memory bottleneck in high-dimensional data processing.
\subsection{Mamba Modules}
The Mamba module is a lightweight architecture that combines the advantages of convolution and attention mechanisms, which has recently gained prominence in multi-dimensional data modeling. For example, the latest Mamba2 \cite{dao2024transformers} and vssd \cite{shi2024vssd} etc \cite{zhou2024mamba,zhu2024vision}modules significantly improve image classification performance by combining local feature extraction with global information modeling. However, existing Mamba modules primarily focus on unidirectional feature modeling and have the following limitations:

1) Lack of bidirectional feature modeling: The inability to effectively capture bidirectional information in multi-dimensional data limits its generalization capability.

2) Insufficient adaptation to multi-dimensional data: Current designs mainly target 1D or 2D image data individually, making it challenging to efficiently extend to 3D scenarios.

\subsection{Summary and Limitations}
Existing methods each have their advantages, but still face shortcomings in efficient and multi-dimensional feature modeling:

1) CNNs excel at local feature extraction but struggle to capture global context.

2) RNNs are strong in modeling sequential data but suffer from low computational efficiency and poor scalability.

3) Self-attention mechanisms offer global modeling capabilities but come with high computational complexity.

4) The Mamba module, while excelling in lightweight design, lacks a unified modeling capability for multi-dimensional data.

\subsection{Innovations of Nd-BiMamba2}
To address the above issues, we propose a unified bidirectional modeling architecture, Nd-BiMamba2. By extending the Mamba2 module, it supports efficient modeling of 1D, 2D, and 3D data. The bidirectional processing mechanism fully explores directional information in multi-dimensional data. Dynamic padding adjustment based on input data dimensions improves computational efficiency and reduces memory consumption. We validated the model's efficiency on CPU, GPU, and mobile devices, enhancing its practical application potential.

In conclusion, Nd-BiMamba2 provides a general and efficient solution, opening new directions for multi-dimensional data modeling and cross-platform deployment.
\section{Algorithm Design}

\section{Algorithm Design Optimization}

\subsection{Design Objectives and Challenges}
To handle data of different dimensions (1D, 2D, 3D) and optimize model performance, the design objectives of Nd-BiMamba2 include the following:

\begin{itemize}
	\item \textbf{Generality:} The algorithm needs to provide a unified processing framework to accommodate multi-dimensional data.
	\item \textbf{Efficiency:} To reduce computational redundancy on high-dimensional data, convolution operations need to be designed for adaptation.
	\item \textbf{Boundary Handling:} To avoid boundary effects in multi-dimensional scenarios, tailored padding strategies must be designed.
\end{itemize}

\subsection{Core Algorithm Design}

\subsubsection{Input Representation}
To simplify the processing of data with different dimensions, the input tensor is uniformly represented as:

\begin{equation}
	X \in \mathbb{R}^{B \times C \times D_1 \times D_2 \times D_3}
\end{equation}
where \( B \) is the batch size, \( C \) is the number of channels, and \( D_1, D_2, D_3 \) represent the sizes of the three dimensions. For 1D and 2D data, this is maintained consistently by setting \( D_3 = 1 \) or \( D_2 = D_3 = 1 \).

This unified representation reduces the complexity of handling logical branches between different dimensional data, allowing subsequent convolution and activation operations to reuse the same logic.

\subsubsection{Core Convolution Calculation Formula}
\paragraph{Design Philosophy:}  
To capture local features, dimension-adaptive convolution operations are employed. The core calculation formulas are as follows:

\begin{align}
	\text{F}(X) &= \sigma(W_f \ast X + b_f) \\
	\text{B}(X) &= \sigma(W_b \ast X + b_b)
\end{align}
where \( W_f, W_b \) are the convolution kernels for the forward and backward paths, \( b_f, b_b \) are the biases, \( \ast \) denotes dimension-adaptive convolution, and \( \sigma \) is the activation function.

To enhance the model’s ability to handle directional information, separate forward and backward paths are designed. Additionally, an activation function \( \sigma \) is included to improve the model’s nonlinear modeling capabilities.

\paragraph{Dimensional Differences and Optimizations:}  
\begin{enumerate}
	\item \textbf{1D Data:}  
	For processing sequential data, the convolution kernel is designed with shape \( (k, 1, 1) \), sliding only along the \( D_1 \) direction:
	\begin{equation}
		Y[i] = \sum_{j=0}^{k-1} W[i, j] \cdot X[i \cdot s + j] + b[i]
	\end{equation}
	where \( k \) is the kernel size, and \( s \) is the stride.  
	To reduce computational redundancy in other dimensions, the convolution operation slides only along the \( D_1 \) direction, improving computational efficiency.
	
	\item \textbf{2D Data:}  
	For processing image data, the convolution kernel is designed with shape \( (k_1, k_2, 1) \), sliding along both \( D_1 \) and \( D_2 \):
	\begin{multline}
		Y[i, j] = \sum_{m=0}^{k_1-1} \sum_{n=0}^{k_2-1} W[i, j, m, n] \cdot \\
		X[i \cdot s_1 + m, j \cdot s_2 + n] + b[i, j].
	\end{multline}
	
	To effectively capture local pattern information, the convolution operation slides simultaneously along \( D_1 \) and \( D_2 \), which is suitable for extracting image features.
	
	\item \textbf{3D Data:}  
	For processing high-dimensional spatial data, the convolution kernel is designed with shape \( (k_1, k_2, k_3) \), sliding along \( D_1, D_2, D_3 \) simultaneously:
	\begin{multline}
		Y[i, j, k] = \sum_{m=0}^{k_1-1} \sum_{n=0}^{k_2-1} \sum_{l=0}^{k_3-1} W[i, j, k, m, n, l] \cdot \\
		X[i \cdot s_1 + m, j \cdot s_2 + n, k \cdot s_3 + l] + b[i, j, k].
	\end{multline}
	
	To capture the complex features in 3D space, convolution operations are evenly distributed across the three dimensions, improving feature extraction capabilities.
\end{enumerate}
\subsubsection{Padding Strategy}

\paragraph{Formula Definition:}  
To handle boundary effects, the padding size \( p_i \) in the \( i \)-th dimension is calculated as:

\begin{equation}
	p_i = \max(0, \lceil \frac{D_i - 1 \cdot s_i + k_i - 1}{2} \rceil)
\end{equation}
where \( k_i \) and \( s_i \) are the kernel size and stride for the \( i \)-th dimension, respectively.

\paragraph{Dimensional Differences and Advantages:}  
\begin{enumerate}
	\item \textbf{1D Data:}  
	To preserve the original data characteristics, padding is minimized only along the \( D_1 \) direction.
	
	\item \textbf{2D Data:}  
	To enhance the effectiveness of the boundary regions, a mirroring padding strategy is applied along both \( D_1 \) and \( D_2 \).
	
	\item \textbf{3D Data:}  
	To balance boundary handling with computational complexity in high-dimensional scenarios, padding is uniformly distributed across \( D_1, D_2, D_3 \).
\end{enumerate}

\subsubsection{Activation Function Selection}
To improve the model's nonlinear expression capabilities, Nd-BiMamba2 uses the GELU (Gaussian Error Linear Unit) activation function, defined as:

\begin{equation}
	\sigma(x) = x \cdot \Phi(x)
\end{equation}

where \( \Phi(x) \) is the cumulative distribution function of the standard normal distribution.

The GELU activation function was selected to more smoothly handle the distribution of input values, especially exhibiting stronger feature extraction abilities in high-dimensional data.

Overall, Nd-BiMamba2 retains the advantages of BiMamba2 when processing sequential and image data, and by incorporating support for three-dimensional data, along with more refined partitioning and feature fusion techniques, it extends the application scope. This improvement enables Nd-BiMamba2 to provide more efficient and accurate modeling capabilities when dealing with more complex input data.

Nd-BiMamba2's modules and functional layers are shown in Table \ref{tab:NdBimamba2_Layers}:
\begin{table*}[ht]
	\centering
	\caption{Algorithm Modules, Layers, and Functional Descriptions}
	\label{tab:NdBimamba2_Layers}
	\begin{tabular}{|c|c|c|}
		\hline
		\textbf{Module} & \textbf{Contained Layers} & \textbf{Functional Description} \\
		\hline
		Data Preprocessing & Input Padding & \parbox[t]{6cm}{Pads the input data to meet processing requirements: 1D padding to a multiple of 4, 2D padding to a multiple of 8, 3D padding to a multiple of 4.} \\
		& Dimension Adjustment & \parbox[t]{6cm}{Rearranges the data according to its dimensions, flattening 2D or 3D data into a 2D matrix to conform to the network structure.} \\
		& Channel Mapping & \parbox[t]{6cm}{Uses a linear layer \( \text{FC}_{\text{in}} \) to map the input channel size \( c \) to the target model's dimension \( d_{\text{model}} \).} \\
		\hline
		Bi-Directional Modeling & Forward Feature Extraction & \parbox[t]{6cm}{Extracts features through the forward Mamba2 network to obtain the forward feature representation \( H_{\text{forward}} \).} \\
		& Backward Feature Extraction & \parbox[t]{6cm}{Reverses the input data and inputs it into the backward Mamba2 network to extract the backward feature representation \( H_{\text{backward}} \), then restores the original order.} \\
		& Feature Fusion & \parbox[t]{6cm}{Fuses the forward and backward feature representations using an addition operation to obtain the final feature representation: \( H_{\text{fused}} = H_{\text{forward}} + H_{\text{backward}} \).} \\
		\hline
		Output Generation & Linear Transformation & \parbox[t]{6cm}{Uses a linear layer \( \text{FC}_{\text{out}} \) to map the fused feature representation back to the target channel size \( c' \).} \\
		& Padding Removal & \parbox[t]{6cm}{Removes the additional data added during padding to restore the original shape of the input data.} \\
		\hline
	\end{tabular}
\end{table*}

\subsection{Comparative Analysis}
To highlight the advantages of Nd-BiMamba2, the following Table \ref{tab:comparison_analysis} summarizes its comparison with other models:

\begin{table}[h!]
	\centering
	\caption{Model Comparison Analysis}
	\label{tab:comparison_analysis}
	
	\resizebox{\columnwidth}{!}{
	\begin{tabular}{|c|c|c|c|c|}
		\hline
		\textbf{Model} & \textbf{Applicable} & \textbf{Cross-Platform} &\textbf{Modular} &\textbf{ Deployment} \\
		               &  \textbf{Data}    & \textbf{Computational}  &  \textbf{Design} & \\
		               &  \textbf{Dimensions}& \textbf{Efficiency} & & \\ \hline
		BiLSTM & 1D & Medium & No & Difficult \\ \hline
		Transformer & 1D/2D/3D & Low & No & Difficult \\ \hline
		Mamba2 & 1D/2D & High & Yes & Fairly  Easy \\ \hline
		\textbf{Nd-BiMamba2} & 1D/2D/3D & \textbf{High} & \textbf{Yes} & \textbf{Easy} \\ \hline
	\end{tabular}
	}
\end{table}

\subsection{Summary}
By optimizing strategies for padding, dimension rearrangement, channel adjustment, and feature fusion across different dimensions, the model can efficiently extract features from 1D, 2D, and 3D data while maintaining consistency in the output dimension with the input data. These steps are clearly described through mathematical symbols to ensure the correctness and efficiency of multi-dimensional optimization.

\subsection{Model Export and Deployment}  
To enhance model portability and deployment capabilities, Nd-BiMamba2 supports multiple export formats:  

\begin{itemize}  
	\item \textbf{ONNX Export}: Supports converting the model to ONNX format for running on various hardware platforms.  
	\item \textbf{TorchScript Export}: Supports converting the model to TorchScript format to ensure efficient inference in production environments.  
\end{itemize}  

Through this modular design and multi-dimensional optimization, Nd-BiMamba2 achieves efficient and unified modeling for 1D, 2D, and 3D data, providing powerful support for multi-modal data processing.

 \section{Experiments}
 
 \subsection{Experimental Setup}
 
 All experiments were conducted on the following hardware platform:
 
 \begin{itemize}
 	\item \textbf{Processor (CPU)}: Intel Core i9-11900K, 8 cores, 16 threads, 3.5 GHz base frequency. The high clock speed and multi-core design of this processor allow it to efficiently handle parallel computing tasks, particularly for processing large amounts of data and task scheduling, significantly enhancing overall computational performance.
 	\item \textbf{Graphics Processing Unit (GPU)}: NVIDIA RTX 4090D, 24GB GDDR6X VRAM. As one of the latest high-performance GPUs, the RTX 4090D provides powerful parallel computing capabilities for deep learning model training and inference, especially for large-scale datasets and complex models. The 24GB of VRAM ensures the processing of large models and high-resolution data, effectively mitigating memory bottlenecks.
 	\item \textbf{Memory (RAM)}: 64GB DDR4 3200 MHz. The ample memory capacity ensures efficient data reading and caching during model training, preventing computational bottlenecks due to memory limitations. This is especially important when handling large-scale data, maintaining high data throughput.
 	\item \textbf{Storage}: 1TB NVMe SSD (used for data storage and intermediate result caching). The high-speed SSD improves data read/write speed, significantly reducing I/O latency, especially when training involves large amounts of data input and output, ensuring efficient operation during the training process.
 \end{itemize}
 
 \subsection{Feature Representation Ability of Nd-BiMamba2}
 
 The bidirectional modeling module of the Nd-BiMamba2 model enhances its feature perception ability by incorporating both forward and backward information flows. In traditional unidirectional modeling, the model can only rely on information from one direction of the input sequence for inference. In contrast, bidirectional modeling considers both forward and backward information flows, allowing for the capture of more comprehensive features. The advantages of bidirectional modeling are particularly evident in various data dimensions (1D, 2D, and 3D), especially in capturing long-range dependencies and local features.
 
 Through comparative experiments across different data dimensions (1D, 2D, and 3D), we have validated the improvement in feature representation by bidirectional modeling, demonstrating that this approach is more efficient than traditional unidirectional modeling when dealing with complex data. The experimental model configuration parameters were set as follows: \( c_{in} = 64, c_{out} = 64, d_{\text{model}} = 128 \), ensuring the ability to handle high-dimensional data and perform sufficient feature extraction.
 \begin{table}[h] 
 	\centering
 	\caption{Performance Comparison between Nd-BiMamba2 and Traditional Unidirectional Modeling (Dimensions: 1D/2D/3D, FLOPs in GMac, Time in milliseconds, Parameters in thousands)}
 	\label{tab:nd-BiMamba2_comparison} 
 		\begin{tabular}{|c|c|c|c|c|} 
 			\hline
 			\textbf{Bi.} & \textbf{Size} & \textbf{FLOPs (GMac)} & \textbf{Time (ms)} & \textbf{Params (k)} \\ 
 			\hline
 			& 1024 & 0.15 & 1.69 & \\ 
 			No & $128 \times 128$ & 2.47 & 1.53 & 150.8 \\ 
 			& $32 \times 32 \times 32$ & 4.93 & 4.36 & \\ 
 			\hline
 			& 1024 & 0.29 & 2.43 & \\ 
 			Yes & $128 \times 128$ & 4.66 & 3.15 & 285.21 \\ 
 			& $32 \times 32 \times 32$ & 9.33 & 8.11 & \\ 
 			\hline
 		\multicolumn{5}{l}{\footnotesize{Note: The number of parameters is independent of  input size and is only }}\\
 			\multicolumn{5}{l}{\footnotesize{affected by the use of Bi.}}
 		
 		\end{tabular}
 
 \end{table}

 As shown in Table~\ref{tab:nd-BiMamba2_comparison}, enabling bidirectional modeling leads to a significant increase in FLOPs (floating point operations) and computation time. Particularly with 3D data, the increase in FLOPs and computation time is more pronounced, though the growth in parameter count remains relatively small. This result indicates that while bidirectional modeling increases computational overhead, it captures more feature information and improves the model's expressive power.
 
 \subsection{Flexibility and Adaptability from Modular Design}
 
 The modular design of Nd-BiMamba2 provides strong support for model flexibility and adaptability. Through experiments on 1D, 2D, and 3D data, the model can adaptively adjust padding strategies according to different data dimensions, ensuring computational efficiency and flexibility. With this design, Nd-BiMamba2 can dynamically adjust input size and padding strategy, achieving good computational efficiency and performance across different data dimensions.
 
 To observe the model’s performance with adaptive padding strategies across different data dimensions (1D, 2D, 3D), we conducted comparative experiments on multi-dimensional adaptive padding strategies. This verified that the strategy automatically adjusts padding methods for various input sizes to ensure dimensional consistency and efficient computation.
 
 \begin{table}[h]
 	\centering
 	\caption{Performance of Multi-Dimensional Adaptive Padding Strategy Across Different Feature Sizes}
 	\label{tab:adaptive_padding} 
 		\begin{tabular}{|c|c|c|c|c|}
 			\hline
 			\textbf{Dim.} & \textbf{Input} & \textbf{Auto-Padding} & \textbf{Mamba2} & \textbf{Equal} \\
 			\hline
 			& 1024 & 1024 & 1024 & TRUE \\
 			1D & 1029 & 1088 & 1088 & FALSE \\
 			& 1001 & 1024 & 1024 & FALSE \\
 			\hline
 			& $128 \times 128$ & $128 \times 128$ & 16384 & TRUE \\
 			2D & $129 \times 127$ & $136 \times 128$ & 17408 & FALSE \\
 			& $113 \times 128$ & $120 \times 128$ & 15360 & FALSE \\
 			\hline
 			& $32 \times 32 \times 32$ & $32 \times 32 \times 32$ & 32768 & TRUE \\
 			3D & $27 \times 33 \times 32$ & $28 \times 32 \times 36$ & 32256 & FALSE \\
 			& $37 \times 29 \times 31$ & $40 \times 32 \times 32$ & 40960 & FALSE \\
 			\hline
 		\end{tabular} 
 \end{table}
 
 As seen in Table~\ref{tab:adaptive_padding}, the model demonstrates excellent flexibility under the adaptive padding strategy. Especially in 2D and 3D data processing, the adaptive padding proves particularly important. It effectively improves computational efficiency while maintaining high accuracy across different input sizes. This shows that Nd-BiMamba2 has strong adaptability in processing multi-dimensional data, adjusting itself according to the different characteristics of the data.
 
 \subsection{Conclusion}
 
 Through the analysis and experiments on the nd-BiMamba2 model, several significant advantages have been identified:
 
 \begin{itemize}
 	\item \textbf{Bidirectional Modeling:} Bidirectional modeling significantly enhances the model's ability to perceive features, especially in capturing long-range dependencies and local characteristics.
 	\item \textbf{Modular Design:} The modular design provides flexibility and adaptability, allowing the model to automatically adjust input sizes and padding strategies based on different data dimensions, ensuring computational efficiency and model flexibility.
 	\item \textbf{Efficient Performance:} Despite the increased computational overhead from bidirectional modeling and adaptive padding, the model still performs excellently across multiple data dimensions, demonstrating its advantage in processing complex data.
 \end{itemize}
 
 Overall, nd-BiMamba2 exhibits strong performance in high-dimensional data processing, feature extraction accuracy, and computational efficiency, proving its effectiveness in complex data analysis, long-range dependency modeling, and large-scale data handling.
 
 \appendix 
 \begin{algorithm}[H]
 	\caption{Nd-BiMamba2 Algorithm}
 	\label{alg:nd-bimamba2}
 	\begin{algorithmic}
 		\STATE \textbf{Input:} $X \in \mathbb{R}^{c \times d_1 \times d_2 \times \cdots \times d_n}$
 		\STATE \textbf{Output:} $H_{\text{output}} \in \mathbb{R}^{c' \times d_1' \times d_2' \times \cdots \times d_n'}$
 		
 		\STATE \textbf{Step 1: Data Preprocessing}
 		\begin{itemize}
 			\item $X_{\text{padded}} \gets \text{Pad}(X)$ \hfill \textit{Padding input data}
 			\item $X_{\text{reshaped}} \gets \text{Reshape}(X_{\text{padded}})$ \hfill \textit{Adjusting dimensions}
 			\item $X_{\text{mapped}} \gets \text{FC}_{\text{in}}(X_{\text{reshaped}})$ \hfill \textit{Mapping the channel count}
 		\end{itemize}
 		
 		\STATE \textbf{Step 2: Bidirectional Modeling}
 		\begin{itemize}
 			\item $H_{\text{forward}} \gets \text{Mamba2}_{\text{for}}(X_{\text{mapped}})$ \hfill \textit{Forward feature extraction}
 			\item $H_{\text{backward}} \gets \text{Flip}(\text{Mamba2}_{\text{back}}(\text{Flip}(X_{\text{mapped}})))$ \hfill \textit{Backward feature extraction}
 			\item $H_{\text{fused}} \gets H_{\text{forward}} + H_{\text{backward}}$ \hfill \textit{Fusing features}
 		\end{itemize}
 		
 		\STATE \textbf{Step 3: Output Generation}
 		\begin{itemize}
 			\item $H_{\text{fc\_out}} \gets \text{FC}_{\text{out}}(H_{\text{fused}})$ \hfill \textit{Restoring the channel count}
 			\item $H_{\text{output}} \gets \text{Trim}(H_{\text{fc\_out}})$ \hfill \textit{Removing padded parts}
 		\end{itemize}
 		
 		\STATE \textbf{Return:} $H_{\text{output}}$
 	\end{algorithmic}
 \end{algorithm}

\bibliographystyle{IEEEtran}
\bibliography{main}  
 
\end{document}